\documentclass[preprint,12pt]{elsarticle}

\usepackage{multirow}
\usepackage{booktabs}
\usepackage{url}
\usepackage{amssymb}
\usepackage{pifont}
\newcommand{\xmark}{\ding{55}}%
\usepackage[export]{adjustbox}
\usepackage[colorlinks=true,linkcolor=red]{hyperref}
\usepackage{amsmath}
\usepackage{cleveref}
\usepackage{algorithm}
\usepackage{algorithmic}

\journal{}

\begin{document}
\begin{frontmatter}
\title{Spatial self-supervised Peak Learning and correlation-based Evaluation of peak picking in Mass Spectrometry Imaging}
\author{Philipp Weigand\corref{cor1}\fnref{label1,label2}}
\ead{p.weigand@th-mannheim.de}
\cortext[cor1]{Corresponding author.}

\author{Nikolas Ebert\fnref{label1}}
\author{Shad A. Mohammed\fnref{label1,label3}}
\author{Denis Abu Sammour\fnref{label1}}
\author{Carsten Hopf\fnref{label1,label3}}
\author{Oliver Wasenmüller\fnref{label1,label2}}

\affiliation[label1]{organization={Research and Transfer Center CeMOS},
            addressline={Technical University of Applied Sciences Mannheim, Paul-Wittsack-Straße 10},
            city={Mannheim},
            postcode={68163},
            country={Germany}}

\affiliation[label2]{organization={Faculty of Biosciences Heidelberg University},
            addressline={Im Neuenheimer Feld 234},
            city={Heidelberg},
            postcode={69120},
            country={Germany}}

\affiliation[label3]{organization={Mannheim Center for Translational Neurosciences (MCTN), Med. Faculty Mannheim, Heidelberg University},
            addressline={Theodor Kutzer-Ufer 1-3},
            city={68167},
            postcode={Mannheim},
            country={Germany}}

\begin{abstract}
Mass spectrometry imaging (MSI) enables label-free visualization of molecular distributions across tissue samples but generates large and complex datasets that require effective peak picking to reduce data size while preserving meaningful biological information. Existing peak picking approaches perform inconsistently across heterogeneous datasets, and their evaluation is often limited to synthetic data or manually selected ion images that do not fully represent real-world challenges in MSI.
To address these limitations, we propose an autoencoder-based spatial self-supervised peak learning neural network that selects spatially structured peaks by learning an attention mask leveraging both spatial and spectral information. We further introduce an evaluation procedure based on expert-annotated segmentation masks, allowing a more representative and spatially grounded assessment of peak picking performance.
We evaluate our approach on four diverse public MSI datasets using our proposed evaluation procedure. Our approach consistently outperforms state-of-the-art peak picking methods by selecting spatially structured peaks, thus demonstrating its efficacy.
These results highlight the value of our spatial self-supervised network in comparison to contemporary state-of-the-art methods. The evaluation procedure can be readily applied to new MSI datasets, thereby providing a consistent and robust framework for the comparison of spatially structured peak picking methods across different datasets.
\end{abstract}

\begin{keyword}
Mass spectrometry imaging \sep Peak picking \sep Quantitative evaluation \sep Self-supervised learning \sep Autoencoder
\end{keyword}

\end{frontmatter}

\section{Introduction}
Mass spectrometry imaging (MSI) enables label-free analysis of molecular distributions for biological tissue analysis. It is used for molecular imaging in pharmaceutical research and drug development \cite{schulz2019advanced}, discovery of diagnostic biomarkers \cite{mechref2012identifying} and histopathological analysis of tissue samples \cite{aichler2015maldi}. Advances in the spatial and mass resolution of contemporary instruments have led to the generation of vast amounts of data from a single measurement. Peak picking is a commonly employed preprocessing step \cite{alexandrov2013testing, wijetunge2015exims, gibb2012maldiquant, bemis2015cardinal} that filters distinct signals while simultaneously removing artifacts and noise from MSI data. When applied properly, peak picking has the crucial advantage of preserving the integrity of the original data, ensuring minimal loss of critical information during preprocessing. Additionally, by reducing data volume, peak picking facilitates subsequent tasks such as spatial segmentation \cite{guo2022isegmsi}, clustering of ion images \cite{zhang2021spatially}, probabilistic metabolite mapping \cite{abu2023spatial} and classification of tissue regions \cite{behrmann2018deep, kanter2023classification}.

The spectra of an MSI dataset can be in either profile or centroided format. Profile spectra retain the full peak shapes and typically contain the same number of m/z values in each pixel. Centroiding, or peak picking, reduces these peak shapes to a single m/z value per peak. Most peak picking methods (see \Cref{tab:sota_peakpicking}) process spectra separately from each other, without spatial context. For example, MALDIquant \cite{gibb2012maldiquant} and Cardinal \cite{bemis2015cardinal} provide a modular analysis pipeline for MSI data, which also includes peak picking. These peak picking approaches usually employ thresholding of signal-to-noise ratios on single spectra, in order to distinguish distinct peaks from noisy signals. However, they do not incorporate the spatial information present in MSI data. Sparse frame multipliers have been introduced as the first approach that utilizes spatial and spectral (spatio-spectral) information for peak picking of structured ion images in matrix-assisted laser desorption/ionization (MALDI) time-of-flight (TOF) MSI datasets. 

\begin{table}[h]
\caption{Overview of existing state-of-the-art peak picking methods \cite{gibb2012maldiquant, lieb2020peak, inglese2019sputnik, abdelmoula2021peak}.\label{tab:sota_peakpicking}}
\resizebox{\textwidth}{!}{
\begin{tabular}{lcccccc}
\toprule
Method name & Input & Functionality & Quantitative evaluation\\
\midrule
MALDIquant \cite{gibb2012maldiquant} / Cardinal \cite{bemis2015cardinal} & Single spectra & Savitzky-Golay-Filter, S/N-Thresholds & \xmark\\
msiPL \cite{abdelmoula2021peak} & Single spectra &  Variational Autoencoder & \xmark \\
Lieb et al. \cite{lieb2020peak} & Spatio-spectral & Sparse frame multipliers & \checkmark \\
SPUTNIK \cite{inglese2019sputnik} & Ion images & Image filters, PCA-image correlation & \xmark \\
\bottomrule
\end{tabular}}
\end{table}

Some approaches focus solely on ion images, neglecting the spectral context of MSI data. For instance, the measure of spatial chaos was developed to assess the spatial structure of ion images \cite{alexandrov2013testing}, and an improved version of Gray level Co-Occurrence (GCO) matrix \cite{gadelmawla2004vision} was proposed for similar purposes \cite{wijetunge2015exims}. A refined measure of spatial chaos was later introduced to evaluate the spatial structure of ion images as part of a pipeline designed to identify biologically relevant molecular signals \cite{palmer2017fdr}. Both the spatial chaos measure and the GCO matrix approach are specifically developed for assessing spatial structure and are evaluated on single ion images only, making them inefficient for peak picking of large profile MSI datasets.

In contrast, SPUTNIK \cite{inglese2019sputnik} is a preprocessing pipeline that also focuses on ion images, but is explicitly designed for peak picking in profile MSI datasets. SPUTNIK computes a reference image for the MSI dataset using principal component analysis (PCA) \cite{pearson1901liii}. Ion images are then correlated with the reference image to identify relevant peaks.

The first deep learning-based peak picking method, msiPL \cite{abdelmoula2021peak}, utilizes a variational autoencoder \cite{kingma2013auto} to reconstruct single spectra. Subsequently, relevant peaks are identified by analyzing the weights of the network. Despite the advancements in MSI preprocessing, to the best of our knowledge, there is currently no deep learning-based peak picking method that effectively utilizes the spatial information inherent in MSI data.

As classical peak picking methods, such as those provided in MALDIquant \cite{gibb2012maldiquant} and Cardinal \cite{bemis2015cardinal}, are designed to detect distinct peaks independently of the spatial structure of the corresponding ion images, we use the term \textit{spatially structured peak picking} for methods that select peaks in profile MSI data by incorporating the spatial structure of peaks within their method. Currently, the approach from Lieb et al. \cite{lieb2020peak} and SPUTNIK \cite{inglese2019sputnik} are designed for spatially structured peak picking. The evaluation of classical peak picking and spatially structured peak picking methods has usually been conducted qualitatively on real-world profile MSI data or quantitatively on simulated data. For instance, the evaluation of mass spectrometry (MS) peak picking has relied on manually chosen distinct, positive peaks as a ground truth \cite{bauer2010evaluation}. Simulation of MALDI-TOF MSI data has been proposed early on to study peak characteristics \cite{coombes2005understanding} and to assess peak picking algorithms against a predefined ground truth \cite{lieb2020peak}. Alternatively, evaluation was conducted on a small expert-curated subset of $200$ spatially structured and $50$ non-structured ion images, which are selected from an MSI dataset comprising $3045$ ion images in total. The subset was used to evaluate several methods which access the spatial structure of ion images \cite{alexandrov2013testing, wijetunge2015exims, palmer2017fdr}. However, none of these evaluation procedures employs a real-world profile MSI dataset for a quantitative evaluation of spatially structured peak picking. Furthermore, the current evaluation procedures are not easily transferable to new datasets without extensive manual annotation.

In this paper, we address the lack of consistent performance of existing peak picking and spatially structured peak picking methods across diverse datasets and the limited evaluation of spatially structured peak picking methods. For this purpose, we introduce spatial self-supervised peak learning, S\textsuperscript{3}PL, an end-to-end trainable neural network for peak picking of spatially structured ion images. Building on the work of Cordes et al. \cite{cordes2024pym2aia}, we leverage the spatio-spectral data screening approach within a deep learning framework to enhance spatially structured peak picking in MSI data. 
Our approach demonstrates superior results compared to state-of-the-art peak picking and spatially structured peak picking methods across three diverse and publicly available MSI datasets.
Moreover, we propose a novel evaluation procedure for spatially structured peak picking methods based on the correlation between ion images and an expert-annotated segmentation mask. The mask captures spatial structures within the MSI data and is utilized to generate a ground truth of spatially structured, positive peaks and non-informative, negative peaks. This approach offers a practical and scalable solution for the evaluation of spatially structured peak picking methods across diverse MSI datasets.


\section{Methods}\label{sec:methods}
In this section, we present our novel method for spatial self-supervised peak learning, S\textsuperscript{3}PL. For the first time, we utilize spatio-spectral processing of MSI data in the context of deep learning-based peak picking for spatially structured ion images. The following sections describe the model architecture, data requirements, and preprocessing techniques, followed by details about the training procedure. As a second contribution, we introduce a novel correlation-based evaluation procedure, which allows for the evaluation of spatially structured peak picking methods on real-world profile MSI datasets, provided that a corresponding expert-annotated segmentation mask is available.

\subsection{Spatial self-supervised peak learning}\label{subsec:autoencoder}
\subsubsection{Model architecture}\label{subsubsec:architecture}
\begin{figure}[t!]%
    \centering
    \includegraphics[width=1.03\linewidth]{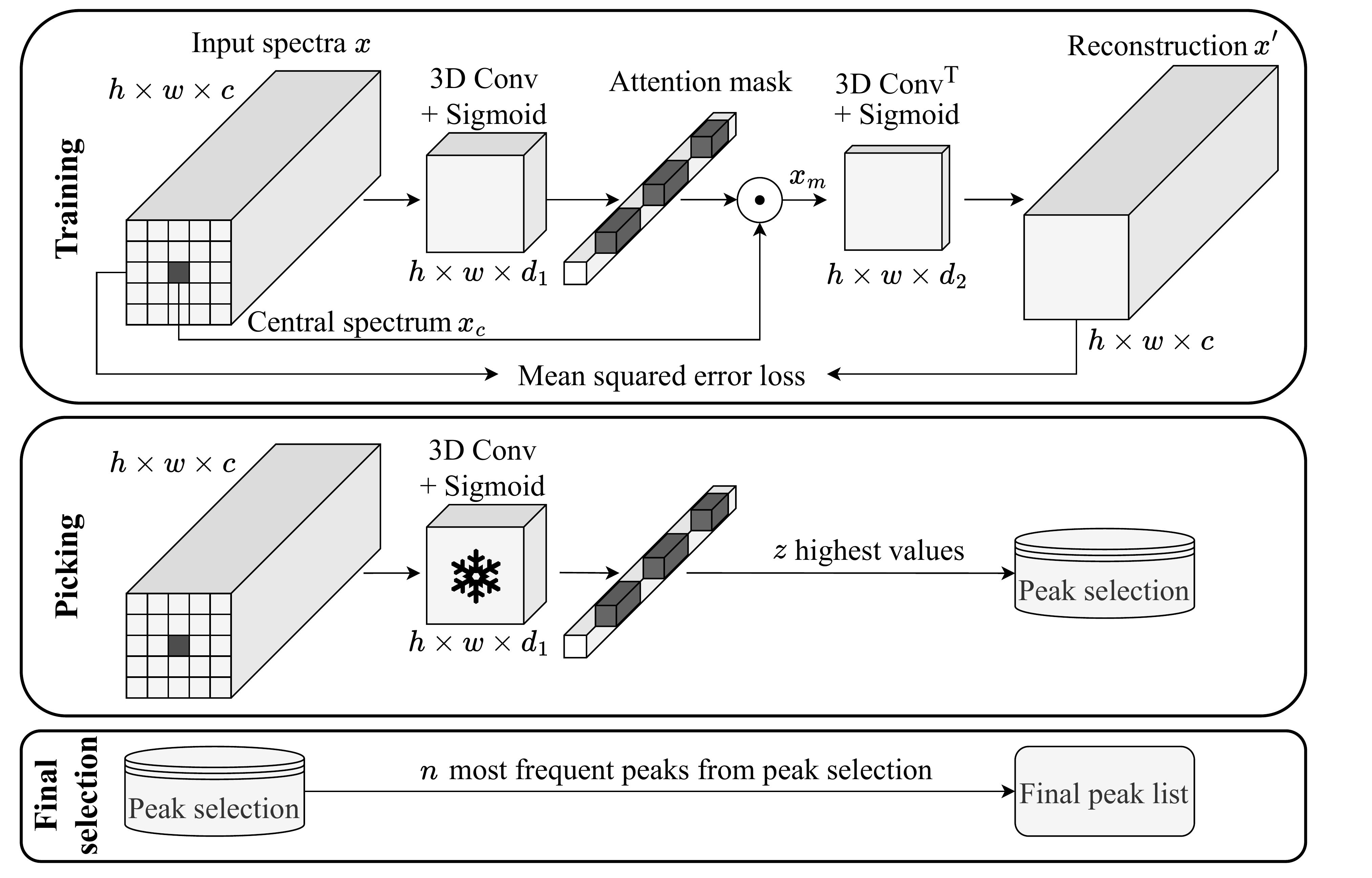}
    \caption{Our proposed spatial self-supervised peak learning pipeline, S\textsuperscript{3}PL, for spatially structured peak picking. The network receives an input comprising a spectral patch $x$ of a complete spectrum and its neighboring spectra. During training, the network learns to identify the most informative peaks by learning an attention mask. In order to identify the peaks, the frozen 3D convolution is applied to the entire dataset, resulting in the filtering of $z$ peaks per spectral patch $x$. In the final selection, the $n$ most frequent peaks of the peak selection yield the final peak list.}
    \label{fig:autoencoder_pipeline}
\end{figure}

Our proposed spatial self-supervised peak learning pipeline, S\textsuperscript{3}PL, is illustrated in \Cref{fig:autoencoder_pipeline}.
For extracting both spatial and spectral features within the spectral patch $x \in \mathbb{R}^{h \times w \times c} $ with a quadratic patch size $p$ resulting in $x \in \mathbb{R}^{p^{2} \times c} $, we utilize a streamlined 3D convolutional autoencoder. In this context, $h$, $w$ and $c$ refer to the height and width of the spectral patch $x$ and the number of $m/z$ values in the MSI data, respectively.
First, the neural network uses a 3D convolution with the kernel-size $h \times w \times d_{1}$, where $d_{1}$ denotes the kernel depth of the 3D convolution, in order to learn a continuous attention mask of size $1 \times 1 \times c$. Similar to approaches from hyperspectral imaging \cite{dou2020band,zhang2024hyperspectral}, the attention mask is then multiplied element-wise with the central spectrum $x_{c} \in \mathbb{R}^{1 \times 1 \times c}$ of the input patch $x$, thereby highlighting informative $m/z$ values for the reconstruction. This operation can be described as
\begin{equation}
x_{m} = \sigma(Conv3D(x)) \odot x_{c}, \label{eq_mask_multiplication}
\end{equation}
where $\sigma$, $\sigma(Conv3D(x))$, $\odot$ and $x_{c}$ refer to the sigmoid function, the attention mask, the hadamard product (element-wise product) and to the central spectrum of the input patch $x$ .
Subsequently, a 3D transposed convolution with the kernel-size $h \times w \times d_{2}$ facilitates the reconstruction $x' \in \mathbb{R}^{p^{2} \times c}$ of the input patch $x$ by transforming the spectrum to the original input shape. Here, $d_{2}$ refers to the kernel depth of the 3D transposed convolution. A sigmoid activation function is applied subsequent to both the 3D convolution and the 3D transposed convolution. The optimal parameters are selected on the basis of the experimental results obtained in \Cref{subsec:ablation_study}. All used parameters are summarized in Supplementary Table S1.

\subsubsection{Preprocessing}\label{subsubsec:preprocessing}
Our proposed neural network is designed for spatially structured peak picking on profile MSI data. We apply the commonly used total ion count (TIC) normalization \cite{norris2007processing} to reduce non-informative spectral variations and batch effects \cite{balluff2021batch}. 
Our model requires the same number of spectral features ($m/z$ values) in each pixel of the MSI dataset. Therefore, binning may be required as an initial step depending on the data. For memory-efficient data handling we utilize pyM\textsuperscript{2}aia \cite{cordes2024pym2aia} to import the files in imzML format \cite{rompp2011imzml}.

\subsubsection{Model training}\label{subsubsec:training}
The model is trained on all spectral patches $X$ in the MSI dataset, so that the number of training samples equals the number of spectra in the dataset. The loss function is designed as the mean squared error (MSE) between the reconstructed and original spectral patch. The ADAM optimizer \cite{kingma2014adam} is employed with a learning rate of $0.01$ and a batch size of $16$. The model is trained on a single NVIDIA GeForce RTX 4070 Ti graphics card for $10$ epochs until the model reaches convergence. The duration of the training process varies between $46$ and $200$ seconds, depending on the size of the dataset, which is primarily influenced by the number of $m/z$ values. The overall training procedure is described in \cref{alg:training}.

\begin{algorithm}
\caption{Training Procedure}\label{alg:training}
\begin{algorithmic}
\REQUIRE Spectral patches $X$, learning rate $\alpha$, number of epochs $T$
\STATE Initialize model parameters $\theta$
\FOR{$epoch = 1$ to $T$}
  \FOR{spectral patch $x$ in all spectral patches $X$}
    \STATE $\text{AttentionMask} \gets \sigma(\text{Conv3D}(x))$
    \STATE $x_m \gets x_c \odot \text{AttentionMask}$
    \STATE $x' \gets \sigma(Conv3D^{T}(x_m))$
    \STATE Compute MSE loss $\mathcal{L} = \|x' - x\|^2$
    \STATE Backpropagate and update weights
  \ENDFOR
\ENDFOR
\end{algorithmic}
\end{algorithm}

\subsubsection{Spatially structured peak picking}\label{subsubsec:peakpicking}
In order to achieve optimal selection of spatially structured peaks, we freeze the previously learned 3D convolution and apply it, together with the sigmoid function, to all spectral patches $X$, with the objective of computing an attention mask for each spectral patch $x$. Unlike during training, the attention mask is not multiplied with the central spectrum $x_{c}$, but analyzed by selecting the $z$ highest activations.
The corresponding $m/z$ values are subsequently added to the peak selection. This process is repeated for each spectral patch $x$, resulting in a peak selection that includes multiple $m/z$ values with varying frequencies. The final peak list is then formed by selecting the $n$ most frequent peaks from the peak selection. The entire peak picking procedure is covered by \cref{alg:peakpicking}.

\begin{algorithm}
\caption{Peak Picking Using Learned Attention Mask}\label{alg:peakpicking}
\begin{algorithmic}
\REQUIRE Spectral patches $X$, desired number of peaks per patch $z$
\STATE $\text{PeakSelection} \gets [\ ]$
\FOR{spectral patch $x$ in all spectral patches $X$}
  \STATE $\text{AttentionMask} \gets \sigma(\text{Conv3D}(x))$ 
  \STATE Extract $\text{top-}z$ m/z-values from the highest $\text{AttentionMask}$ activation
  \STATE Append extracted m/z-values to $\text{PeakSelection}$
\ENDFOR
\STATE Determine the $n$ most frequently occurring m/z peaks in $\text{PeakSelection}$
\end{algorithmic}
\end{algorithm}

\subsection{Pearson correlation-based evaluation of spatially structured peak picking methods}\label{subsec:pearson_evaluation}

\begin{figure}[t!]
    \centering
    \includegraphics[width=1\linewidth]{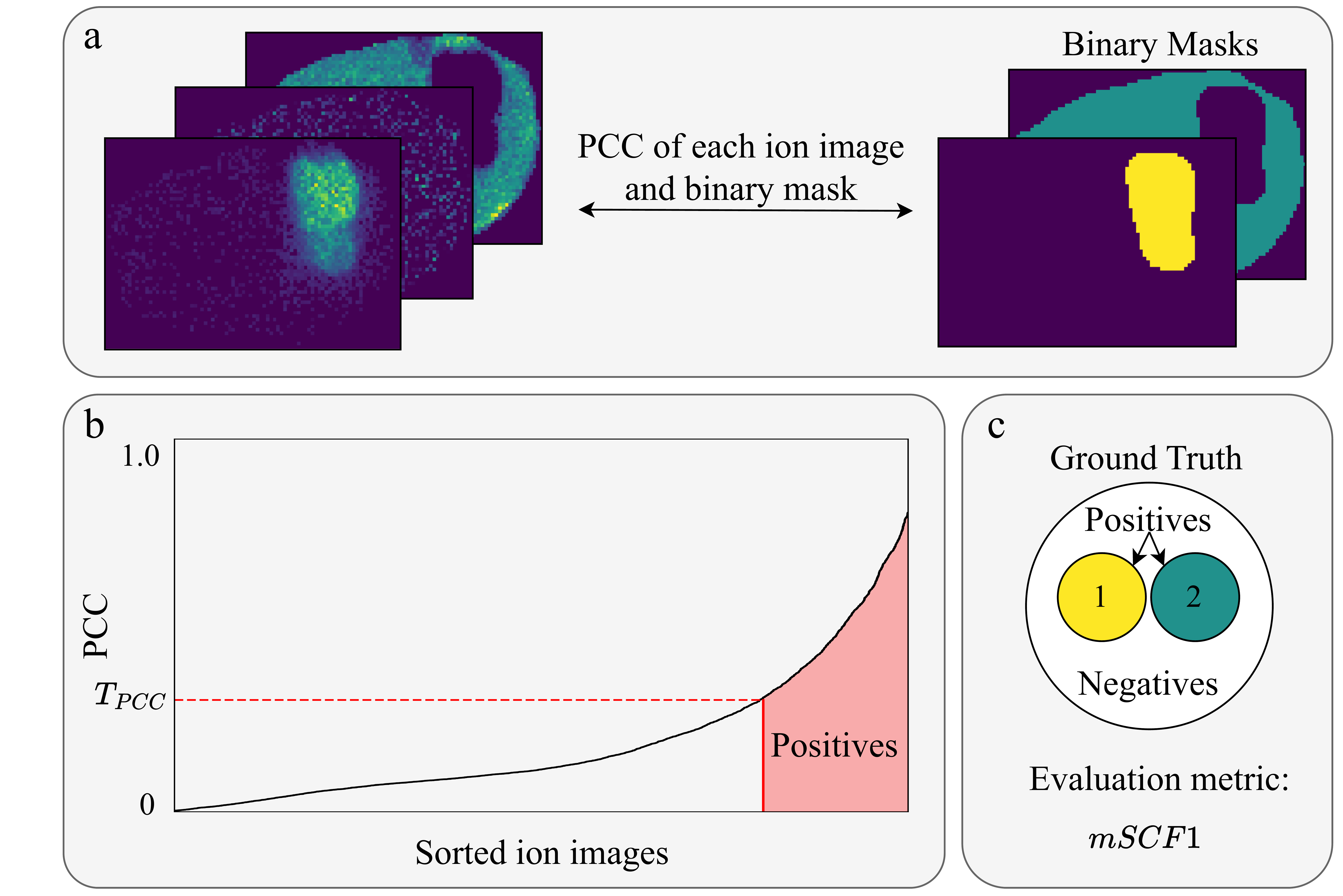}
    \caption{Our proposed evaluation procedure for peak picking of spatially structured peaks. (a) PCC values of all ion images and a corresponding binary mask are computed and sorted. (b) All ion images with a PCC value equal to or greater than $T_{PCC}$ for any annotated class are classified as positive. Steps (a) and (b) are repeated for every class. (c) We calculate F1-scores using thresholds $T_{PCC} \in \{0.3, 0.4, 0.5, 0.6\}$ and report the mean of these four F1-scores, $mSCF1$, as the final metric.}
    \label{fig:pcc_explanation}
\end{figure}

The spatial information of ion images is important, as it forms the basis for visualization of biomolecular distributions and for spatial probabilistic mapping \cite{abu2023spatial}. Previous peak picking evaluation procedures for MSI data do not always use real-world profile datasets quantitative evaluation. Instead, visual inspection is conducted for qualitative assessment \cite{lieb2020peak} and simulated or preselected MSI data is employed for quantitative evaluation \cite{lieb2020peak, alexandrov2013testing}. Hence, we propose an evaluation procedure (see \Cref{fig:pcc_explanation}) for real-world profile MSI datasets based on the spatial structure of ion images. Our evaluation procedure utilizes an expert-annotated segmentation mask as a reference for spatial structures within the MSI data. The mask serves as the definition of the ground truth for the relevant peaks to be picked. 
The positive ground truth, i.e., the peaks that should be identified by a peak picking method, is determined by selecting all ion images that exhibit a high degree of correlation with any structures indicated in the segmentation mask.
The degree of correlation between an ion image and a binary mask is quantified using the Pearson correlation coefficient (PCC), which has been demonstrated to be an effective metric for assessing correlations between ion images in a previous study \cite{ovchinnikova2020colocml}. The PCC has also been successfully applied to correlate ion images with binary masks generated by segmentation algorithms \cite{abdelmoula2021peak, abdelmoula2022massnet}. The ion images are not further processed before applying the PCC, as we do not want to introduce further parameters that would entail an additional source of bias in the ground truth generation process. The PCC yields the same result when intensity scalings like normalization are applied to the ion images, due to its invariance to linear transformations. For empty or near-constant ion images, the PCC is not defined due to a standard deviation of zero. In such cases, a PCC of zero is assigned. In our work, we suggest that the PCC should be used to correlate ion images with expert-annotated segmentation masks.

We define a threshold $T_{PCC}$ that differentiates between ion images that are informative (positives) and those that are non-informative (negatives). An ion image with a PCC value equal to or greater than $T_{PCC}$ for any annotated structure in the segmentation mask is classified as positive. Ion images with a PCC value below $T_{PCC}$ for all structures in the segmentation mask are classified as negatives. With the ground truth established, we calculate the F1-score in accordance with previous publications \cite{lieb2020peak, yang2009comparison} as
\begin{equation}
F1_{T_{PCC}}=\frac{2TP}{2TP+FP+FN},\label{eq_F1}
\end{equation}
where $TP$, $FP$ and $FN$ denote the true positives, false positives and false negatives, respectively.

The PCC threshold $T_{PCC}$ serves to differentiate between structured and non-structured ion images by setting a fixed value. The choice of this threshold is crucial, as it directly impacts the evaluation outcomes. Furthermore, relying on a single threshold can introduce a bias. To address this issue, we propose combining multiple thresholds $T_{PCC} \in \{0.3, 0.4, 0.5, 0.6\}$ by averaging their $F1_{T_{PCC}}$ results as
\begin{equation}
mSCF1=\frac{F1_{0.3}+F1_{0.4}+F1_{0.5}+F1_{0.6}}{4},\label{eq_F1_mean}
\end{equation}
yielding the final metric for our evaluation procedure: the \textit{mean spatial correlation F1-score} ($mSCF1$). This approach allows us to incorporate a range of strict to lenient thresholds, thus broadening the scope of our analysis. By doing so, we reduce the influence of subjective threshold selection and make the evaluation procedure more robust. Furthermore, this method enables a fair comparison of different techniques that may target various aspects of image structure. In other words, we ensure that no method is unfairly advantaged or disadvantaged by the choice of a single threshold. Further threshold parameters are analysed in Supplementary Figures S1-S3.


\section{Results}\label{sec:results}

To demonstrate the general applicability of our spatially structured peak picking method S\textsuperscript{3}PL, we conduct a comparative analysis with state-of-the-art peak picking methods \cite{gibb2012maldiquant, lieb2020peak, inglese2019sputnik, abdelmoula2021peak} (see \Cref{tab:sota_peakpicking}), using three publicly available dataset collections (see \Cref{tab:datasets} and Supplementary Table S2 for tissue specific information). First, we present the public datasets that we utilize for the evaluation of spatially structured peak picking using our proposed Pearson correlation-based evaluation procedure from \Cref{subsec:pearson_evaluation}. We then compare our results to those of the state-of-the-art methods. Furthermore, we conduct ablation studies to determine the best hyperparameters for our S\textsuperscript{3}PL method.

\subsection{Datasets}
In order to demonstrate the broad applicability of our S\textsuperscript{3}PL method, we employ three diverse and publicly accessible datasets \cite{abdelmoula2022massnet, bemis2019cardinalworkflows, inglese2017deep} (see \Cref{tab:datasets}) for quantitative evaluation. To the best of our knowledge, we selected all publicly available datasets that meet the necessary criteria for our evaluation procedure, namely those containing profile MSI data along with expert-annotated segmentation masks. In total, eight tissue sections are used for each of the three dataset collections in the quantitative evaluation, resulting in a total of $24$ tissue sections. Subsequently, the expert-annotated segmentation masks corresponding to each tissue section in the dataset collections are utilized for evaluation purposes. 
In order to show generalization over different mass analyzers, we also include a MALDI-TOF dataset \cite{abu2019quantitative}. A quantitative evaluation was not feasible for this sample because the provided segmentation mask does not correspond to the structures visible in the ion images. The dataset expresses subtle regions which have not been annotated by the histopathologist and would therefore lead to a misleading ground truth with artificially low mSCF1 scores. Therefore, this tissue is used for qualitative comparison (Figure \ref{fig:ionimages_of_picked_peaks}).

\begin{table}[t]
\caption{Overview of public dataset collections \cite{abdelmoula2022massnet, bemis2019cardinalworkflows, inglese2017deep, abu2019quantitative} used for the evaluation of spatially structured peak picking methods.
\label{tab:datasets}}
\resizebox{\textwidth}{!}{
\begin{tabular}{lcccccc}
\toprule%
Dataset & Format & \# $m/z$ values & \# tissues &  \# classes & Ionization & Analyzer\\
\midrule
Mouse glioblastoma \cite{abdelmoula2022massnet} & Profile & 85,062 & 8 & 2 & MALDI & FTICR \\
Renal cell carcinoma \cite{bemis2019cardinalworkflows}& Profile & 10,200 & 8 & 3 & DESI & Thermo LTQ Orbitrap \\
Colorectal adenocarcinoma \cite{inglese2017deep} & Profile & 1,481 & 8 & 3 & DESI & Thermo Exactive \\
Gastrointestinal stromal tumor \cite{abu2019quantitative} & Profile & 120,000 & 1 & - & MALDI & TOF \\
\bottomrule
\end{tabular}}
\end{table}

The first MSI dataset collection, which was previously published, uses a patient-derived xenograft (PDX) mouse brain tumor model of glioblastoma (GBM) \cite{abdelmoula2022massnet} comprising eight GBM tissue sections, each containing annotated tumor and non-tumor regions. As the data set lacks off-sample spectra, a background class has not been annotated. The MALDI-MSI measurement was conducted using a 9.4 Tesla SolariX Fourier transform ion cyclotron resonance (FTICR) mass spectrometer (Bruker Daltonics, Billerica, MA) in the positive ion mode with a spatial resolution of $100$ \textmu m and a mass range of $100$ to $1,000$ resulting in $85,062$ $m/z$ values.

The renal cell carcinoma (RCC) cancer dataset from \textit{CardinalWorkflows} \cite{bemis2019cardinalworkflows} is the second dataset collection and comprises eight pairs of human kidney tissue. Each tissue pair consists of a normal tissue sample and a cancerous tissue sample, resulting in the annotated classes \textit{tumor}, \textit{healthy tissue} and \textit{background}. The MSI data for each tissue pair was acquired using a desorption electrospray ionization (DESI) on a Thermo Finnigan LTQ ion trap mass spectrometer in negative ion mode. The mass range covered $150$ to $1,000$ with $10,200$ $m/z$ values.

The third dataset collection utilizes a biopsy of human colorectal adenocarcinoma (CAC) \cite{inglese2017deep}.
The DESI-MSI data was acquired using a Thermo Fisher Exactive mass spectrometer, in the negative ion mode within a mass range of $200$ to $1,050$, yielding $1,481$ $m/z$ values. For this data, hematoxylin and eosin (H\&E) tissue staining images are available and annotated by a histopathologist into three categories; \textit{tumor}, \textit{healthy tissue} and \textit{background}.
We selected eight of the $52$ tissue sections in order to avoid the inclusion of visibly duplicate sections. Tissue sections were selected from each region of the biopsy to represent the entire specimen. Specifically, the selected tissue sections were \textit{40TopL}, \textit{160TopL}, \textit{200TopL}, \textit{240TopL}, \textit{280TopL}, \textit{360TopL}, \textit{400TopL} and \textit{520TopL}, which correspond to the top-left tissue in the imzML files. As pseudo ground truth segmentation masks, we use the $99.9\%$ accurate predictions of a deep learning model \cite{inglese2017deep}, because the corresponding expert-annotated masks are not available.

The fourth dataset \cite{abu2019quantitative} contains MALDI-TOF-MSI spectra of a gastrointestinal stromal tumor (GIST) tissue acquired using an ultrafleXtreme instrument in positive ion mode within a mass range of $300$ to $2,000$, resulting in $120,000$ $m/z$ values. We chose the tissue sample \textit{GIST\_sampleA\_tumor\_TOF} for analysis, as it contains both tumor and normal tissue, which may present a challenge due to the heterogeneity of the tissue.

\subsection{Comparison to the State of the Art}\label{subsec:comparison}
In all state-of-the-art peak picking methods, the number of detected peaks depends on the specific parameters used. To ensure a fair comparison across methods, we define the number of peaks to be selected for each tissue section based on our established ground truth from \Cref{subsec:pearson_evaluation}. For each dataset collection, we select a PCC threshold that results in a reasonable number of ground truth peaks for the tissue sections in that dataset. This defined number of peaks to be picked is then used consistently across all methods. 
The chosen PCC thresholds are 0.4, 0.5 and 0.3 for the GBM, RCC and CAC datasets, respectively. Different PCC thresholds and therefore different number of peaks are necessary for each dataset, because the PCC values for the best-correlating ion images vary significantly across the dataset collections, leading to large discrepancies in the number of peaks to be picked. For instance, the RCC dataset contains numerous ion images with PCC values exceeding 0.5 for all classes, while the CAC dataset has very few ion images above this threshold.

The parameters of all state-of-the-art methods are individually adjusted for each tissue section, thereby enabling us to approximate the number of picked peaks to a value that is close to the defined number of peaks to be picked for each tissue section. More precisely, we perform a grid search of the parameter corresponding to the method until the number of picked peaks results in the range of ground truth peaks $\pm 50$. In some cases, this was not possible, because even the smallest or highest possible parameters did not result in the corresponding peak number range. The used parameters for each state-of-the-art method are listed in Supplementary Table S3 and the tested parameters for the grid search are shown in Supplementary Tables S4-S7. For our S\textsuperscript{3}PL method, we select the exact number of peaks to be picked by using the intended parameter for the number of peaks $n$. We provide a sensitivity analysis of the $mSCF1$ scores of our S\textsuperscript{3}PL method in Supplementary Table S8. Further details regarding parameter optimization for our method are addressed in the ablation study in \Cref{subsec:ablation_study}.

The Cardinal package \cite{bemis2015cardinal} is not used in this work. Instead, the MALDIquant package \cite{gibb2012maldiquant} is employed, as both packages utilize similar techniques for peak picking. Lieb et al.'s method \cite{lieb2020peak} produces a smoothed spectrum, wherein every signal exceeding zero is considered as a peak. This results in significantly more peaks than ground truth positives, leading to poor results. Note that, unlike the original method of Lieb et al., we introduce a threshold for their method, considering only m/z values that exceed the threshold as peaks, leading to more accurate results. We use this threshold parameter to optimize the results of Lieb et al.'s method for the evaluation.

\begin{figure}[h!]%
    \centering
    \includegraphics[width=1\linewidth]{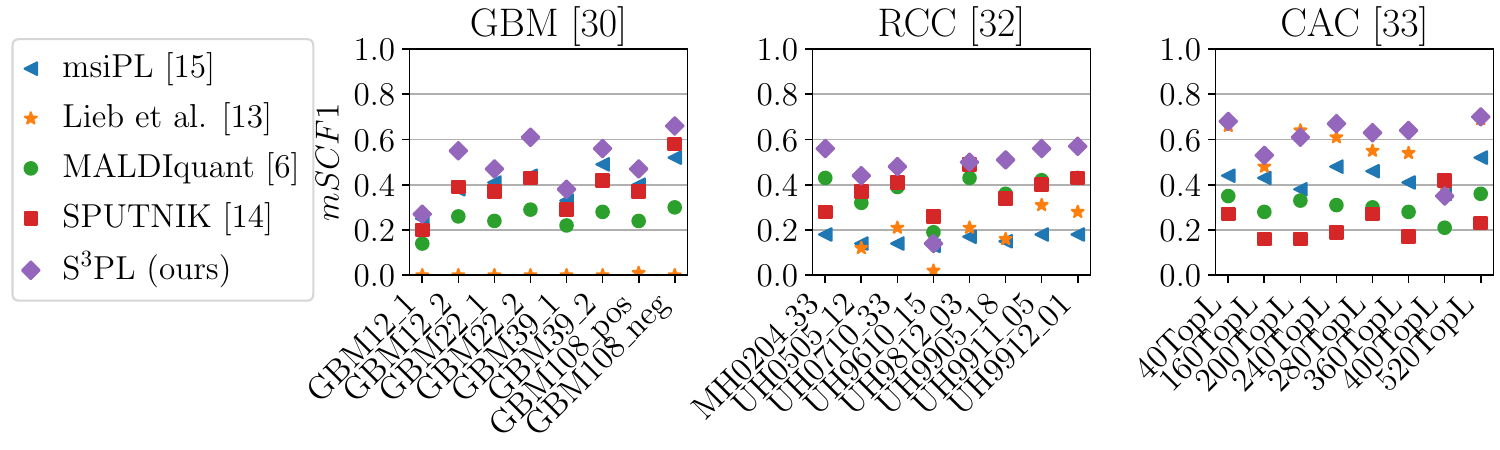}
    \caption{Detailed comparison of our spatial self-supervised peak learning autoencoder, S\textsuperscript{3}PL, with state-of-the-art peak picking methods \cite{gibb2012maldiquant, lieb2020peak, inglese2019sputnik, abdelmoula2021peak} using our proposed evaluation procedure. The figures show the $mSCF1$ scores for each tissue section within each dataset separately.}
    \label{fig:result_figure}
\end{figure}

In \Cref{tab:results}, we report multiple F1-scores, which are averaged over all tissue sections in each dataset. We use $T_{PCC}$ values of $0.3$, $0.4$, $0.5$, and $0.6$, along with their mean metric, $mSCF1$. Detailed results of $mSCF1$-scores for each tissue section are presented in \Cref{fig:result_figure}. Our S\textsuperscript{3}PL method yields the best results across all F1 metrics on the GBM dataset \cite{abdelmoula2022massnet}. In comparison to the second-best result achieved by msiPL \cite{abdelmoula2021peak}, our method demonstrates an improvement of $+9.3\%$ in terms of $mSCF1$.
On the RCC dataset \cite{bemis2019cardinalworkflows}, S\textsuperscript{3}PL achieves the highest scores across all F1 metrics with an improvement of $+9.9\%$ in terms of $mSCF1$ compared to the second-best result by MALDIquant \cite{gibb2012maldiquant}. On the CAC dataset \cite{inglese2017deep}, S\textsuperscript{3}PL yields the highest F1-scores with an improvement of $+11.3\%$ for $mSCF1$ compared to the second-best result by Lieb et al. 

\begin{table*}[t]
\caption{Comparison of our spatial self-supervised peak learning autoencoder, S\textsuperscript{3}PL, with state-of-the-art peak picking methods \cite{gibb2012maldiquant, lieb2020peak, inglese2019sputnik, abdelmoula2021peak} using our proposed evaluation procedure. The table shows the F1-scores for $T_{PCC} \in \{0.3, 0.4, 0.5, 0.6\}$ and the mean of all four F1-scores, $mSCF1$. We average the results over all tissue sections in a dataset. Reproducibility Note: We introduce and optimize a threshold for Lieb et al.'s method, considering only m/z
values that exceed the threshold as peaks, because this leads to more accurate results.
\label{tab:results}}
\resizebox{\textwidth}{!}{
\begin{tabular}{ccccccc}
\hline
\textbf{Dataset}     & \textbf{Metric}                        & \textbf{msiPL \cite{abdelmoula2021peak}}           & \textbf{Lieb et al. \cite{lieb2020peak}}     & \textbf{MALDIquant \cite{gibb2012maldiquant}}      & \textbf{SPUTNIK \cite{inglese2019sputnik}}         & \textbf{S\textsuperscript{3}PL (Ours)}            \\ \hline
\multirow{6}{*}{GBM \cite{abdelmoula2022massnet}} & F1\textsubscript{0.3} & 45.6                     & 0                       & 29.3                     & \underline{50.8}                     & \textbf{56.4}                     \\
                     & F1\textsubscript{0.4} & \underline{44.4}                     & 0                        & 26.8                     & 43.0                     & \textbf{55.6}                     \\
                     & F1\textsubscript{0.5} & \underline{39.4}                     & 0                        & 23.6                     & 34.4                     & \textbf{48.8}                     \\
                     & F1\textsubscript{0.6} & \underline{31.9}                     & 0                        & 18.5                     & 23.8                     & \textbf{37.1}                     \\
                     & mSCF1                                  & \underline{40.3} & 0                        & 24.6 & 38.1 & \textbf{49.6} \\
                     & 95\% CI of mSCF1                                 & {[}33.0, 47.5{]} & -                       & {[}20.4, 28.9{]} & {[}28.9, 47.3{]} & {[}39.0, 60.2{]} \\ \hline
\multirow{6}{*}{RCC \cite{bemis2019cardinalworkflows}} & F1\textsubscript{0.3} & 17.1                     & 32.4                     & 41.4                     & \underline{46.5}                     & \textbf{48.0}                     \\
                     & F1\textsubscript{0.4} & 16.9                     & 27.6                     & 41.6                     & \underline{45.4}                     & \textbf{51.9}                     \\
                     & F1\textsubscript{0.5} & 15.9                     & 21.5                     & \underline{36.8}                     & 34.0                     & \textbf{48.9}                     \\
                     & F1\textsubscript{0.6} & 12.9                     & 13.8                     & \underline{27.9}                     & 22.3                     & \textbf{38.6}                     \\
                     & mSCF1                    & 15.9 & 23.8 & \underline{37.1} & 37.0 & \textbf{47.0} \\
                     & 95\% CI of mSCF1         & {[}14.1, 17.6{]} & {[}11.8, 28.0{]} & {[}30.2, 44.1{]} & {[}30.8, 43.7{]} & {[}35.2, 58.8{]} \\ \hline
\multirow{6}{*}{CAC \cite{inglese2017deep}} & F1\textsubscript{0.3} & 56.3                     & \underline{72.3}                     & 38.4                     & 28.6                     & \textbf{77.9}                     \\
                     & F1\textsubscript{0.4} & 50.1                     & \underline{67.9}                     & 34.8                     & 25.4                     & \textbf{72.1}                     \\
                     & F1\textsubscript{0.5} & 39.9                     & \underline{54.5}                     & 28.1                     & 22.2                     & \textbf{56.0}                     \\
                     & F1\textsubscript{0.6} & 26.1                     & \underline{34.0}                     & 18.0                     & 16.6                     & \textbf{34.5}                     \\
                     & mSCF1 & 43.1 & \underline{57.1} & 29.9 & 23.3 & \textbf{60.1} \\
                    & 95\% CI of mSCF1 & {[}39.1, 47.9{]} & {[}48.9, 65.3{]} & {[}26.3, 34.2{]} & {[}16.0, 30.7{]} & {[}50.6, 69.7{]} \\ \hline
\end{tabular}}
\end{table*}

In \Cref{fig:result_figure}, the $mSCF1$ results for our S\textsuperscript{3}PL method and state-of-the-art methods are shown separately for each tissue section. S\textsuperscript{3}PL outperforms all state-of-the-art methods on each tissue section of the GBM dataset. On the RCC dataset, we outperform the state of the art on every tissue section, except for the \textit{UH9610\_15} tissue section, where SPUTNIK achieves the best $mSCF1$ score with an eleven percent point improvement over our S\textsuperscript{3}PL method. On the CAC dataset, we achieve the best results on six out of eight tissue sections. On tissue section \textit{200TopL}, Lieb et al. achieves the best $mSCF1$ score being three points better than our S\textsuperscript{3}PL method. On tissue section \textit{400TopL}, S\textsuperscript{3}PL ranks as the fourth best. SPUTNIK achieves the highest $mSCF1$ score, outperforming S\textsuperscript{3}PL by seven percentage points. A detailed list of all F1-scores on all tissue sections is listed in Supplementary Table S9. Separate class results, i.e. only one class is considered as positive, are presented in Supplementary Tables S10-S14. We observed no meaningful variability across repeated runs with different random seeds for our S\textsuperscript{3}PL method.

\begin{figure}[t!]
    \centering
    \includegraphics[width=1\columnwidth,center]{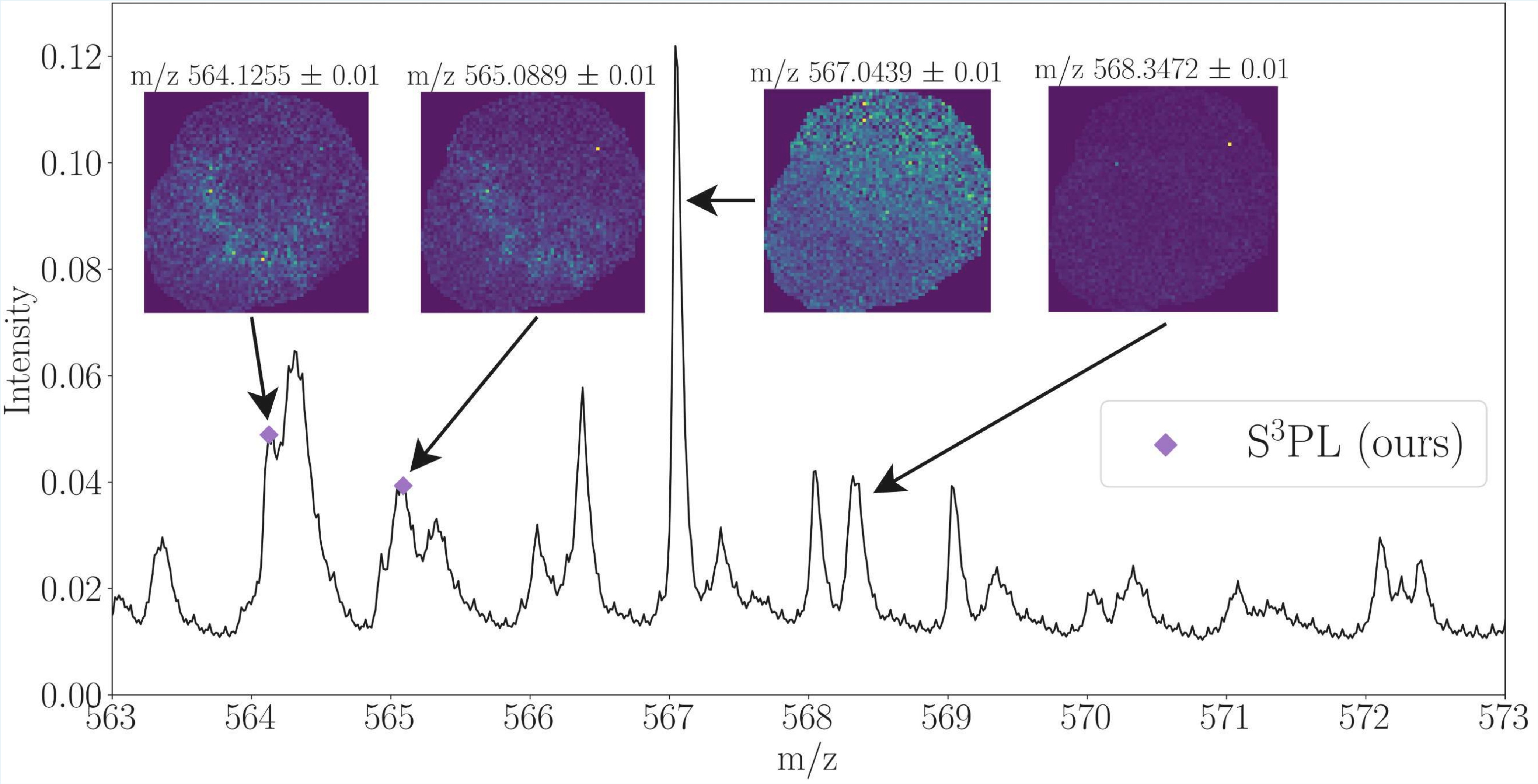}
    \caption{Visualization of selected peaks by our S\textsuperscript{3}PL method on the GIST dataset \cite{abu2019quantitative} in the representative mass range $m/z$ $563$ to $573$. The depicted mean spectrum contains many distinct peaks, which do not necessarily correspond to spatially structured ion images. S\textsuperscript{3}PL only selects peaks with spatial structure.}
    \label{fig:ionimages_of_picked_peaks}
\end{figure}

For the GIST dataset \cite{abu2019quantitative}, we assess the performance of spatially structured peak picking qualitatively by visualizing selected ion images in the range of $m/z$ $563$ to $573$ (\Cref{fig:ionimages_of_picked_peaks} and Supplementary Figures S4 - S8). The classic peak picking method from MALDIquant, which uses signal-to-noise thresholding, and msiPL select many distinct peaks, which do not necessarily exhibit a spatial structure. Therefore, these methods are susceptible to spatial outliers such as isolated high-intensity pixels, which can be misinterpreted as meaningful peaks despite their lack of biological relevance. MALDIquant and msiPL selected six and eight peaks, respectively. SPUTNIK, Lieb et al. and S\textsuperscript{3}PL, which are designed for spatially structured peak picking, select one, two and two ion images in the representative mass range, respectively. Compared to MALDIquant and msiPL, they do not select the peaks with high intensity and uninformative structure. Importantly, S\textsuperscript{3}PL and Lieb et al. select only peaks which correspond to spatially structured ion images.

\subsection{Ablation study}\label{subsec:ablation_study}
In this section, we study the impact of the most important hyperparameters of our model architecture. The objective of the experiments is to identify the optimal hyperparameters, which are the squared spectral patch size $p$, where $p=h=w$, the kernel depths $d_{1}$ and $d_{2}$, the number of highest peaks $z$ to pick per spectral patch as well as the final peak number $n$. 
We use all three dataset collections for the study and average the $mSCF1$ results over all tissue sections. Our S\textsuperscript{3}PL model is trained as described in \Cref{subsubsec:training}. We set the hyperparameters of our S\textsuperscript{3}PL method to the best values being determined by the corresponding experiment in this section.

\begin{figure}[htbp]
    \centering
    \begin{minipage}{0.48\textwidth}
        \centering
        \includegraphics[width=1\linewidth]{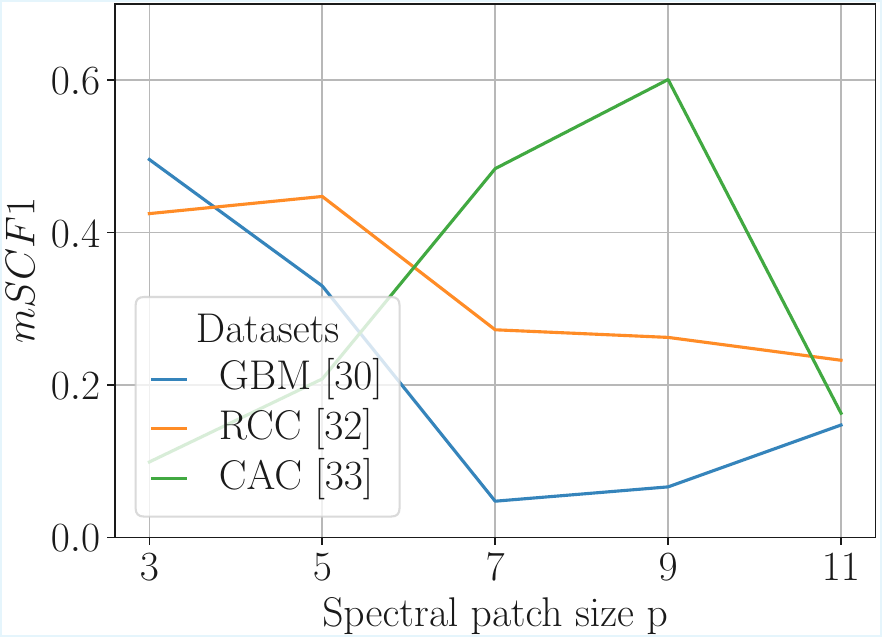}
        \caption{Ablation study on the spectral patch size $p$ across three public datasets \cite{abdelmoula2022massnet, bemis2019cardinalworkflows, inglese2017deep}. The $mSCF1$ score is determined by averaging the results of all tissue sections in a dataset. The best patch size highly depends on the dataset \cite{abdelmoula2022massnet, bemis2019cardinalworkflows, inglese2017deep}.}
    \label{fig:ablation_patch_size}
    \end{minipage}
    \hfill
    \begin{minipage}{0.48\textwidth}
        \centering
         \includegraphics[width=1\linewidth]{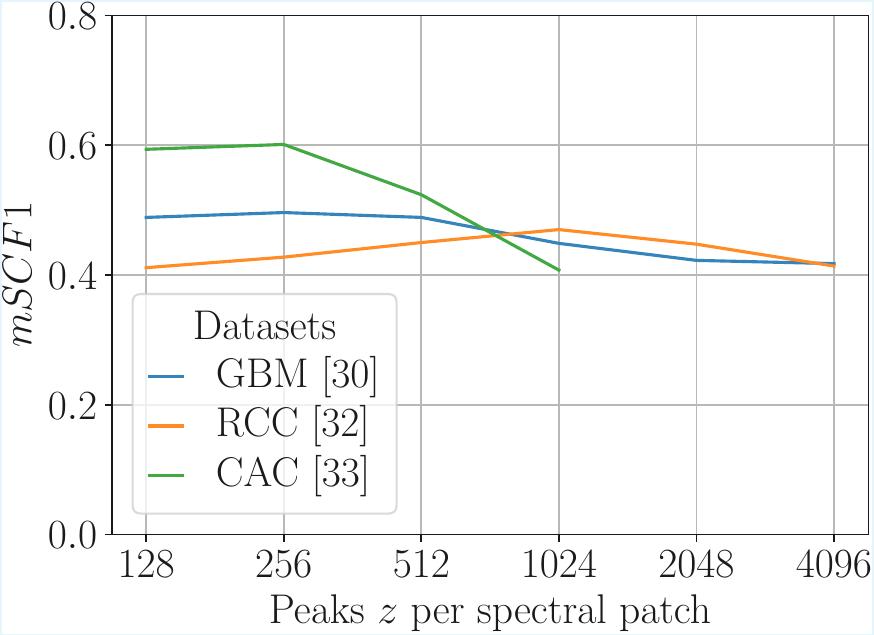}
    \caption{Ablation study on the number of peaks $z$ to pick for every spectral patch, using three public datasets \cite{abdelmoula2022massnet, bemis2019cardinalworkflows, inglese2017deep}. In the case of the CAC dataset \cite{inglese2017deep}, there can not be picked more peaks than $m/z$ values present in the dataset (1481).}
    \label{fig:ablation_number_peaks_per_spectrum}
    \end{minipage}
\end{figure}

\Cref{fig:ablation_patch_size} shows the impact of the spectral patch size $p$ regarding the $mSCF1$ score. For the GBM dataset, best results are achieved with a spectral patch size of $p=3$. For the CAC dataset, the $mSCF1$ score increases with the spectral patch size and has its optimum at $p=9$. For the RCC dataset, the highest $mSCF1$ result is achieved with low spectral patch sizes, the optimum being at $p=5$.

Furthermore, we analyse the best number of peaks $z$ to select for every spectral patch in \Cref{fig:ablation_number_peaks_per_spectrum}. For the GBM and CAC datasets, $z=256$ achieves the best $mSCF1$ results, while the optimum for the RCC dataset is at $z=1024$. The $mSCF1$ score declines for the CAC dataset as $z$ approaches 1481, because this is the total number of $m/z$ values present within the dataset.

\begin{figure}
    \centering
    \includegraphics[width=1.02\linewidth, height=5cm]{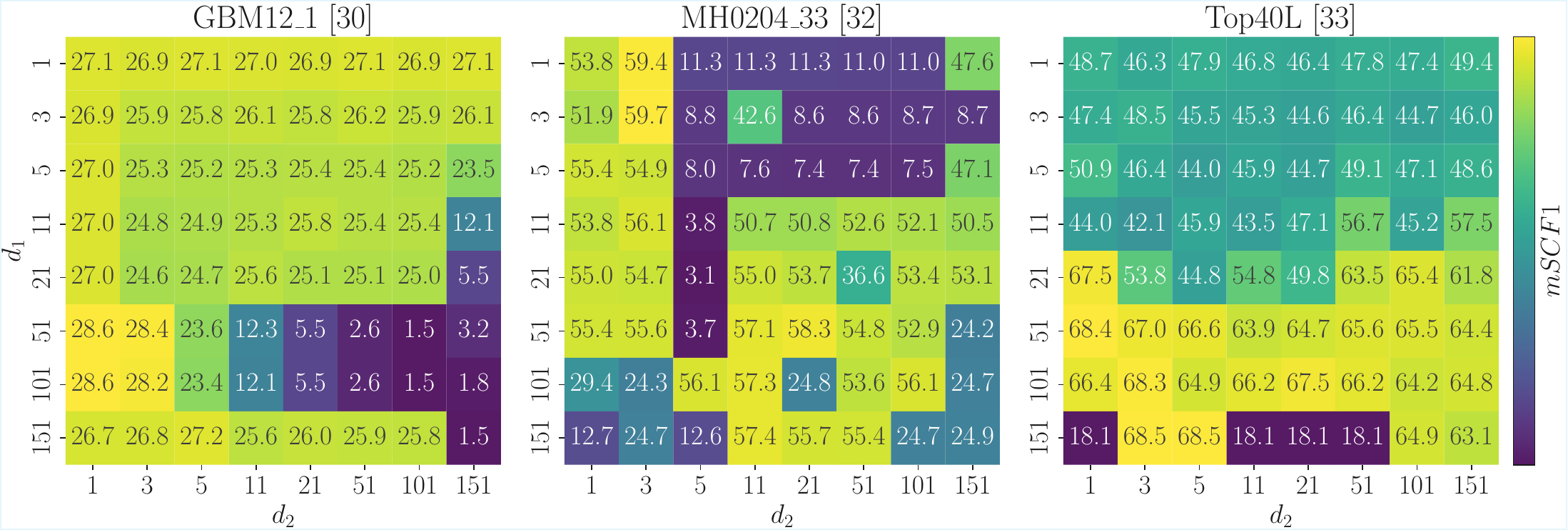}
    \caption{Ablation study on the best kernel depth $d_{1}$ and $d_{2}$. We selected one tissue section from each of the three public datasets \cite{abdelmoula2022massnet, bemis2019cardinalworkflows, inglese2017deep}. The heatmap shows the $mSCF1$ score for different combinations of kernel depths.}
    \label{fig:ablation_kernel_size_depths}
\end{figure}

In order to choose the best kernel depths $d_{1}$ and $d_{2}$, we compare different combinations of both values. \Cref{fig:ablation_kernel_size_depths} shows a heatmap for the first tissue section of each dataset collection. The $mSCF1$ scores are displayed for kernel depths of $1$, $3$, $5$, $11$, $21$, $51$, $101$ and $151$. For all three tissue sections, there is a slight trend of higher $mSCF1$ scores for an increasing kernel depth $d_{1}$. The lower the kernel depth $d_{2}$, the better the $mSCF1$ results. For tissue section \textit{MH0204\_33}, best $mSCF1$ results can also be achieved with certain combinations of high kernel depths. We choose the kernel depth $d_{1}=51$ for the 3D convolution and the kernel depth $d_{2}=1$ for the 3D transposed convolution, because this combination achieves approximately the best $mSCF1$ results over all three tissue sections.
A summary of all used parameters for our S\textsuperscript{3}PL method for each dataset is given in Supplementary Table S1. The number of peaks $n$ is tissue section dependent and is equal to the number of ground truth peaks. These values are listed in Supplementary Table S3.

\begin{figure}[!t]%
    \centering
    \includegraphics[width=1\linewidth]{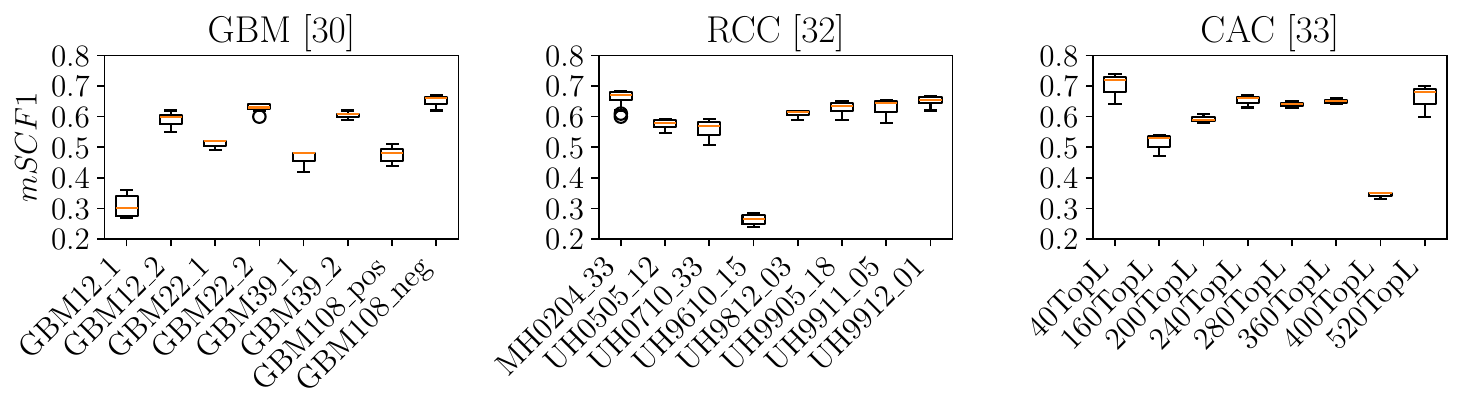}
    \caption{Variation of $mSCF1$ scores with a fine-tuned peak number $n$ on other tissue sections of a dataset collection \cite{abdelmoula2022massnet, bemis2019cardinalworkflows, inglese2017deep}. The boxplots indicate the best and worst possible results when fine-tuning the number of peaks on one tissue section and using this number for every other tissue section of a dataset collection.}
    \label{fig:ablation_peak_tuning}
\end{figure}

The number of peaks to be picked varies across different tissue sections within the dataset collections. In our final experiment, we investigate how well our method S\textsuperscript{3}PL performs, when the final number of peaks $n$ is determined on a single tissue section of a dataset collection and then retained for the other tissue sections of the same dataset collection. In \Cref{fig:ablation_peak_tuning} we depict the boxplots of eight $mSCF1$ scores for each tissue section. The $mSCF1$ scores result from using the best peak numbers of other tissue sections of the corresponding dataset collection, which achieve the best $mSCF1$ scores on these tissue sections. The best peak number for the tissue section itself is also included. Note, that the number of peaks $n$ that achieves the best $mSCF1$ score on a tissue section is not necessarily the number of peaks to be picked for that tissue section. For the GBM tissue sections, the best and worst results differ by a maximum of $9\%$ points for the tissue section \textit{GBM12\_1}, while the smallest deviations of $3\%$ points are observed for tissue section \textit{GBM39\_2}. For the RCC dataset, the biggest deviations occur for the tissue section \textit{UH0710\_33} ($8\%$). For the CAC dataset, maximum deviations of $10\%$ points are observed for the tissue sections \textit{40TopL} and \textit{520TopL}.

\section{Discussion}
In this paper, we presented our S\textsuperscript{3}PL model, which is a spatial self-supervised peak learning neural network for spatially structured peak picking on MSI data. 
S\textsuperscript{3}PL builds upon prior work in spatial processing of MSI data, such as the method by Lieb et al. \cite{lieb2020peak}, and leverages recent advances in self-supervised learning for MSI \cite{abdelmoula2021peak}. While some approaches, e.g. contrastive learning methods for ion image clustering \cite{hu2022self}, also employ self-supervision, they are not designed for peak picking. In contrast, S\textsuperscript{3}PL integrates spatial priors and self-supervised training tailored to the peak picking task. Additionally, our use of a learned attention mask enables the model to selectively emphasize spatial features, contributing to its improved performance over existing approaches. We demonstrated the performance compared to existing state-of-the-art methods \cite{gibb2012maldiquant, lieb2020peak, inglese2019sputnik, abdelmoula2021peak} by yielding superior $mSCF1$ results on three diverse datasets. Previous methods have encountered difficulties in achieving consistent, optimal results on real-world profile MSI datasets with varying numbers of $m/z$ values. Our results demonstrate, that S\textsuperscript{3}PL is a valuable tool for peak picking on any kind of MSI data, regardless of the ion source and the number of $m/z$ values. It is noteworthy that the limited number of weights in our model facilitates rapid training and inference time when compared to neural networks with millions of parameters, such as msiPL \cite{abdelmoula2021peak}.

Furthermore, we introduced a novel evaluation procedure for peak picking methods based on the correlation of ion images to an expert-annotated segmentation mask. Our approach does not require the single assessment of all ion images separately, thereby reducing the labelling effort. Importantly, our evaluation procedure can be transferred to any MSI dataset, given a segmentation mask containing the spatial structures in the data. By using multiple threshold values $T_{PCC} \in \{0.3, 0.4, 0.5, 0.6\}$, we incorporate rather weakly correlating ion images as well as only strong correlating ones to the expert-annotated segmentation mask. Thus, we ensure a fair evaluation with a broad range of analysis. Additionally, multiple threshold values have the advantage that datasets with generally weakly correlating ion images can be analysed. As for any ground truth generation, the segmentation mask must be created with precision, as incomplete or missing regions result in erroneous evaluation outcomes and false interpretation. Our evaluation approach is less affected by individual incorrect pixel annotations, because the mask as a whole is crucial for the correlation with ion images. However, as the generation of a segmentation masks can be challenging especially for heterogeneous tissues, our evaluation metric has to be employed carefully for datasets with limited knowledge about the underlying data. In such cases, a visual inspection as performed on the GIST dataset \cite{abu2019quantitative} is required.

The final number of peaks $n$ to be selected is not automatically chosen by our network, thereby allowing for flexibility in tailoring the selection based on the needs for the specific application. While this flexibility is beneficial, it may present a challenge for users who are unfamiliar with selecting an appropriate number of peaks for their analyzed sample. This challenge is not unique to our method, as other state-of-the-art approaches similarly require user-defined parameters. Automating the selection of $n$ is non-trivial because the attention mask in our model scores m/z values relative to each other, and these scores vary across datasets. As a result, determining a universally optimal number of peaks for unseen datasets remains an open problem. However, we demonstrate in \Cref{fig:ablation_peak_tuning} that selecting an appropriate peak number on one representative tissue section and applying it across the remaining sections of the dataset collection leads to a minimal performance drop, typically less than $10\%$ in mSCF1 for only a few tissue sections, thus supporting the robustness of our approach.

One of the limitations of our S\textsuperscript{3}PL model lies in the requirement for profile data with consistent m/z bins. Furthermore, certain parameters need to be adjusted. Specifically the spectral patch size $p$ and the number of peaks $z$ per spectral patch, need to be adjusted according to the specific dataset being used. Fixed values for these parameters do not consistently achieve optimal results across all datasets. Consequently, we maintain flexibility for the spectral patch size $p$ and the number of peaks $z$ per spectral patch, akin to the approach taken by other state-of-the-art methods.

Future experiments should compare the existing peak picking methods across more datasets. The automatic selection of the number of peaks should be the next focus, with the objective of advancing towards a fully automated pipeline. Spatially structured peak picking could be extended to centroided MSI datasets as a filter for spatially structured peaks. In addition, it should be analyzed, whether peak picking helps to improve the results of post-processing tasks such as classification, clustering, or segmentation on MSI datasets. Unfortunately, the number of publicly available real-world profile MSI datasets, which contain an expert-annotated segmentation mask, is very limited. Hence, other authors are encouraged to upload real-world profile data together with an expert-annotated segmentation mask to facilitate advancements in the MSI field.

\section{Conclusion}
In this paper, we proposed S\textsuperscript{3}PL, the first self-supervised peak learning neural network that incorporates spatial information for spatially structured peak picking in MSI. Compared to state-of-the-art peak picking methods, we achieve superior results on three diverse datasets. In addition, we proposed an evaluation procedure for spatially structured peak picking using a segmentation mask. This provides a consistent and robust approach for comparing spatially structured peak picking methods quantitatively across different profile MSI datasets.

\section*{Declaration of competing interest }
The authors declare that they have no known competing financial
interests or personal relationships that could have appeared to influence the work reported in this paper. 

\section*{CRediT authorship contribution statement}
\textbf{Philipp Weigand:} Conceptualization, Methodology, Software, Writing - Original Draft, Visualization. 
\textbf{Nikolas Ebert: } Methodology, Formal analysis, Writing - Review \& Editing. 
\textbf{Shad A. Mohammed: } Formal analysis, Investigation, Writing - Review \& Editing. 
\textbf{Denis Abu Sammour: } Formal analysis, Resources, Writing - Review \& Editing. 
\textbf{Carsten Hopf: } Conceptualization, Formal analysis, Resources, Writing - Review \& Editing. 
\textbf{Oliver Wasenmüller: } Conceptualization, Methodology, Resources, Writing - Review \& Editing, Supervision.

\section*{Funding}
This project has been funded by the Ministry of Science, Research and Arts Baden-Württemberg [BW6-07 to P.W.];
the Deutsche Forschungsgemeinschaft (DFG, German Research Foundation)
    [Project-ID INST874/9-1 to C.H and O.W,
    Project-ID 404521405, SFB 1389—UNITE Glioblastoma to D.A.S and C.H];
and the Deutscher Akademischer Austauschdienst (DAAD, German Academic Exchange Service) to S.A.M.

\section*{Data availability}
Our source Code is available on GitHub at \url{https://github.com/CeMOS-IS/S3PL}.
The GBM dataset \cite{abdelmoula2022massnet} is available in the NIH Common Fund’s National Metabolomics Data Repository (NMDR) Metabolomics Workbench under project id (PR001292). The RCC dataset is accessible via the R package \textit{CardinalWorkflows} \cite{bemis2019cardinalworkflows}. The CAC dataset \cite{inglese2017deep} can be obtained from the MetaboLights repository under project id (MTBLS415). The GIST tissue section \textit{GIST\_sampleA\_tumor\_TOF} \cite{abu2019quantitative} is available on METASPACE \cite{palmer2017fdr} (\url{https://metaspace2020.eu/}).

\bibliographystyle{elsarticle-num} 
\bibliography{reference}

@article{mechref2012identifying,
  title={Identifying cancer biomarkers by mass spectrometry-based glycomics},
  author={Mechref, Yehia and Hu, Yunli and Garcia, Aldo and Hussein, Ahmed},
  journal={Electrophoresis},
  volume={33},
  number={12},
  pages={1755--1767},
  year={2012},
  publisher={Wiley Online Library}
}

@article{zhang2021spatially,
  title={Spatially aware clustering of ion images in mass spectrometry imaging data using deep learning},
  author={Zhang, Wanqiu and Claesen, Marc and Moerman, Thomas and Groseclose, M Reid and Waelkens, Etienne and De Moor, Bart and Verbeeck, Nico},
  journal={Anal. Bioanal. Chem.},
  volume={413},
  pages={2803--2819},
  year={2021},
  publisher={Springer}
}

@article{guo2022isegmsi,
  title={iSegMSI: an interactive strategy to improve spatial segmentation of mass spectrometry imaging data},
  author={Guo, Lei and Liu, Xingxing and Zhao, Chao and Hu, Zhenxing and Xu, Xiangnan and Cheng, Kian-Kai and Zhou, Peng and Xiao, Yu and Shah, Mudassir and Xu, Jingjing and others},
  journal={Anal. Chem.},
  volume={94},
  number={42},
  pages={14522--14529},
  year={2022},
  publisher={ACS Publications}
}

@article{abdelmoula2021peak,
  title={Peak learning of mass spectrometry imaging data using artificial neural networks},
  author={Abdelmoula, Walid M and Lopez, Begona Gimenez-Cassina and Randall, Elizabeth C and Kapur, Tina and Sarkaria, Jann N and White, Forest M and Agar, Jeffrey N and Wells, William M and Agar, Nathalie YR},
  journal={Nat. Commun.},
  volume={12},
  number={1},
  pages={5544},
  year={2021},
  publisher={Nature Publishing Group UK London}
}

@article{abdelmoula2022massnet,
  title={massNet: integrated processing and classification of spatially resolved mass spectrometry data using deep learning for rapid tumor delineation},
  author={Abdelmoula, Walid M and Stopka, Sylwia A and Randall, Elizabeth C and Regan, Michael and Agar, Jeffrey N and Sarkaria, Jann N and Wells, William M and Kapur, Tina and Agar, Nathalie YR},
  journal={Bioinformatics},
  volume={38},
  number={7},
  pages={2015--2021},
  year={2022},
  publisher={Oxford University Press}
}

@article{lieb2020peak,
  title={Peak detection for MALDI mass spectrometry imaging data using sparse frame multipliers},
  author={Lieb, Florian and Boskamp, Tobias and Stark, Hans-Georg},
  journal={J. Proteomics},
  volume={225},
  pages={103852},
  year={2020},
  publisher={Elsevier}
}

@article{palmer2017fdr,
  title={FDR-controlled metabolite annotation for high-resolution imaging mass spectrometry},
  author={Palmer, Andrew and Phapale, Prasad and Chernyavsky, Ilya and Lavigne, Regis and Fay, Dominik and Tarasov, Artem and Kovalev, Vitaly and Fuchser, Jens and Nikolenko, Sergey and Pineau, Charles and others},
  journal={Nat. Methods},
  volume={14},
  number={1},
  pages={57--60},
  year={2017},
  publisher={Nature Publishing Group US New York}
}

@article{gibb2012maldiquant,
  title={MALDIquant: a versatile R package for the analysis of mass spectrometry data},
  author={Gibb, Sebastian and Strimmer, Korbinian},
  journal={Bioinformatics},
  volume={28},
  number={17},
  pages={2270--2271},
  year={2012},
  publisher={Oxford University Press}
}

@article{bemis2015cardinal,
  title={Cardinal: an R package for statistical analysis of mass spectrometry-based imaging experiments},
  author={Bemis, Kyle D and Harry, April and Eberlin, Livia S and Ferreira, Christina and van de Ven, Stephanie M and Mallick, Parag and Stolowitz, Mark and Vitek, Olga},
  journal={Bioinformatics},
  volume={31},
  number={14},
  pages={2418--2420},
  year={2015},
  publisher={Oxford University Press}
}

@incollection{bauer2010evaluation,
  title={Evaluation of peak-picking algorithms for protein mass spectrometry},
  author={Bauer, Chris and Cramer, Rainer and Schuchhardt, Johannes},
  booktitle={Data Mining in Proteomics: From Standards to Applications},
  pages={341--352},
  year={2010},
  publisher={Springer}
}

@article{alexandrov2013testing,
  title={Testing for presence of known and unknown molecules in imaging mass spectrometry},
  author={Alexandrov, Theodore and Bartels, Andreas},
  journal={Bioinformatics},
  volume={29},
  number={18},
  pages={2335--2342},
  year={2013},
  publisher={Oxford University Press}
}

@article{wijetunge2015exims,
  title={EXIMS: an improved data analysis pipeline based on a new peak picking method for EXploring Imaging Mass Spectrometry data},
  author={Wijetunge, Chalini D and Saeed, Isaam and Boughton, Berin A and Spraggins, Jeffrey M and Caprioli, Richard M and Bacic, Antony and Roessner, Ute and Halgamuge, Saman K},
  journal={Bioinformatics},
  volume={31},
  number={19},
  pages={3198--3206},
  year={2015},
  publisher={Oxford University Press}
}

@article{kingma2013auto,
  title={Auto-encoding variational bayes},
  author={Kingma, Diederik P},
  journal={arXiv preprint arXiv:1312.6114},
  year={2013}
}

@article{kingma2014adam,
  title={Adam: A method for stochastic optimization},
  author={Kingma, Diederik P and Ba, Jimmy},
  journal={arXiv preprint arXiv:1412.6980},
  year={2014}
}

@article{pearson1901liii,
  title={LIII. On lines and planes of closest fit to systems of points in space},
  author={Pearson, Karl},
  journal={The London, Edinburgh, and Dublin philosophical magazine and journal of science},
  volume={2},
  number={11},
  pages={559--572},
  year={1901},
  publisher={Taylor \& Francis}
}

@article{bemis2019cardinalworkflows,
  title={CardinalWorkflows: Datasets and workflows for the Cardinal mass spectrometry imaging package},
  author={Bemis, KA},
  journal={R package version},
  volume={1},
  number={0},
  year={2019}
}

@article{inglese2017deep,
  title={Deep learning and 3D-DESI imaging reveal the hidden metabolic heterogeneity of cancer},
  author={Inglese, Paolo and McKenzie, James S and Mroz, Anna and Kinross, James and Veselkov, Kirill and Holmes, Elaine and Takats, Zoltan and Nicholson, Jeremy K and Glen, Robert C},
  journal={Chem. Sci.},
  volume={8},
  number={5},
  pages={3500--3511},
  year={2017},
  publisher={Royal Society of Chemistry}
}

@article{cordes2024pym2aia,
  title={pyM2aia: Python interface for mass spectrometry imaging with focus on deep learning},
  author={Cordes, Jonas and Enzlein, Thomas and Hopf, Carsten and Wolf, Ivo},
  journal={Bioinformatics},
  volume={40},
  number={3},
  pages={btae133},
  year={2024},
  publisher={Oxford University Press}
}

@article{rompp2011imzml,
  title={imzML: Imaging Mass Spectrometry Markup Language: A common data format for mass spectrometry imaging},
  author={R{\"o}mpp, Andreas and Schramm, Thorsten and Hester, Alfons and Klinkert, Ivo and Both, Jean-Pierre and Heeren, Ron MA and St{\"o}ckli, Markus and Spengler, Bernhard},
  journal={Data Mining in Proteomics: From Standards to Applications},
  pages={205--224},
  year={2011},
  publisher={Springer}
}

@article{inglese2019sputnik,
  title={SPUTNIK: an R package for filtering of spatially related peaks in mass spectrometry imaging data},
  author={Inglese, Paolo and Correia, Gon{\c{c}}alo and Takats, Zoltan and Nicholson, Jeremy K and Glen, Robert C},
  journal={Bioinformatics},
  volume={35},
  number={1},
  pages={178--180},
  year={2019},
  publisher={Oxford University Press}
}

@article{coombes2005understanding,
  title={Understanding the characteristics of mass spectrometry data through the use of simulation},
  author={Coombes, Kevin R and Koomen, John M and Baggerly, Keith A and Morris, Jeffrey S and Kobayashi, Ryuji},
  journal={Cancer Inform.},
  volume={1},
  pages={117693510500100103},
  year={2005},
  publisher={SAGE Publications Sage UK: London, England}
}

@article{zhang2024hyperspectral,
  title={A hyperspectral band selection method based on sparse band attention network for maize seed variety identification},
  author={Zhang, Liu and Wei, Yaoguang and Liu, Jincun and Wu, Jianwei and An, Dong},
  journal={Expert Syst. Appl.},
  volume={238},
  pages={122273},
  year={2024},
  publisher={Elsevier}
}

@article{dou2020band,
  title={Band selection of hyperspectral images using attention-based autoencoders},
  author={Dou, Zeyang and Gao, Kun and Zhang, Xiaodian and Wang, Hong and Han, Lu},
  journal={IEEE Geosci. Remote. Sens. Lett.},
  volume={18},
  number={1},
  pages={147--151},
  year={2020},
  publisher={IEEE}
}

@article{behrmann2018deep,
  title={Deep learning for tumor classification in imaging mass spectrometry},
  author={Behrmann, Jens and Etmann, Christian and Boskamp, Tobias and Casadonte, Rita and Kriegsmann, J{\"o}rg and Maa$\beta$, Peter},
  journal={Bioinformatics},
  volume={34},
  number={7},
  pages={1215--1223},
  year={2018},
  publisher={Oxford University Press}
}

@article{norris2007processing,
  title={Processing MALDI mass spectra to improve mass spectral direct tissue analysis},
  author={Norris, Jeremy L and Cornett, Dale S and Mobley, James A and Andersson, Malin and Seeley, Erin H and Chaurand, Pierre and Caprioli, Richard M},
  journal={Int. J. Mass Spectrom.},
  volume={260},
  number={2-3},
  pages={212--221},
  year={2007},
  publisher={Elsevier}
}

@article{yang2009comparison,
  title={Comparison of public peak detection algorithms for MALDI mass spectrometry data analysis},
  author={Yang, Chao and He, Zengyou and Yu, Weichuan},
  journal={BMC Bioinform.},
  volume={10},
  pages={1--13},
  year={2009},
  publisher={Springer}
}

@article{gadelmawla2004vision,
  title={A vision system for surface roughness characterization using the gray level co-occurrence matrix},
  author={Gadelmawla, ES},
  journal={NDT \& e International},
  volume={37},
  number={7},
  pages={577--588},
  year={2004},
  publisher={Elsevier}
}

@article{ovchinnikova2020colocml,
  title={ColocML: machine learning quantifies co-localization between mass spectrometry images},
  author={Ovchinnikova, Katja and Stuart, Lachlan and Rakhlin, Alexander and Nikolenko, Sergey and Alexandrov, Theodore},
  journal={Bioinformatics},
  volume={36},
  number={10},
  pages={3215--3224},
  year={2020},
  publisher={Oxford University Press}
}

@article{schulz2019advanced,
  title={Advanced MALDI mass spectrometry imaging in pharmaceutical research and drug development},
  author={Schulz, Sandra and Becker, Michael and Groseclose, M Reid and Schadt, Simone and Hopf, Carsten},
  journal={Curr. Opin. Biotechnol.},
  volume={55},
  pages={51--59},
  year={2019},
  publisher={Elsevier}
}

@article{aichler2015maldi,
  title={MALDI Imaging mass spectrometry: current frontiers and perspectives in pathology research and practice},
  author={Aichler, Michaela and Walch, Axel},
  journal={Lab. Invest.},
  volume={95},
  number={4},
  pages={422--431},
  year={2015},
  publisher={Elsevier}
}

@article{balluff2021batch,
  title={Batch effects in MALDI mass spectrometry imaging},
  author={Balluff, Benjamin and Hopf, Carsten and Porta Siegel, Tiffany and Grabsch, Heike I and Heeren, Ron MA},
  journal={J. Am. Soc. Mass Spectrom.},
  volume={32},
  number={3},
  pages={628--635},
  year={2021},
  publisher={ACS Publications}
}

@article{abu2023spatial,
  title={Spatial probabilistic mapping of metabolite ensembles in mass spectrometry imaging},
  author={Abu Sammour, Denis and Cairns, James L and Boskamp, Tobias and Marsching, Christian and Kessler, Tobias and Ramallo Guevara, Carina and Panitz, Verena and Sadik, Ahmed and Cordes, Jonas and Schmidt, Stefan and others},
  journal={Nat. Commun.},
  volume={14},
  number={1},
  pages={1823},
  year={2023},
  publisher={Nature Publishing Group UK London}
}

@article{kanter2023classification,
  title={Classification of Pancreatic Ductal Adenocarcinoma Using MALDI Mass Spectrometry Imaging Combined with Neural Networks},
  author={Kanter, Frederic and Lellmann, Jan and Thiele, Herbert and Kalloger, Steve and Schaeffer, David F and Wellmann, Axel and Klein, Oliver},
  journal={Cancers},
  volume={15},
  number={3},
  pages={686},
  year={2023},
  publisher={MDPI}
}

@article{abu2019quantitative,
  title={Quantitative mass spectrometry imaging reveals mutation status-independent lack of imatinib in liver metastases of gastrointestinal stromal tumors},
  author={Abu Sammour, Denis and Marsching, Christian and Geisel, Alexander and Erich, Katrin and Schulz, Sandra and Ramallo Guevara, Carina and Rabe, Jan-Hinrich and Marx, Alexander and Findeisen, Peter and Hohenberger, Peter and others},
  journal={Sci. Rep.},
  volume={9},
  number={1},
  pages={10698},
  year={2019},
  publisher={Nature Publishing Group UK London}
}

@article{hu2022self,
  title={Self-supervised clustering of mass spectrometry imaging data using contrastive learning},
  author={Hu, Hang and Bindu, Jyothsna Padmakumar and Laskin, Julia},
  journal={Chem. Sci.},
  volume={13},
  number={1},
  pages={90--98},
  year={2022},
  publisher={Royal Society of Chemistry}
}

\end{document}